\documentclass[runningheads]{llncs}
\usepackage[T1]{fontenc}
\usepackage{subcaption}
\usepackage{cite}
\usepackage{footnote}

\usepackage{amssymb}
\usepackage{makecell}
\usepackage{colortbl}
\usepackage{booktabs}
\usepackage[svgnames,table]{xcolor}
\usepackage{amsmath}  
\usepackage{hyperref}

\usepackage{graphicx}
\usepackage{multirow}

\usepackage{rotating}

\usepackage[mathcal]{euscript}

\newcommand*\rot{\rotatebox{90}}

\def\ccb{\cellcolor{blue!20}}
\def\ccB{\cellcolor{blue!20}\textbf}

\usepackage{orcidlink}

\begin{document}
\title{Nearest-Class Mean and Logits Agreement for Wildlife Open-Set Recognition}
\titlerunning{NCMAgreement for Wildlife Open-Set Recognition}
%
\author{Jiahao Huo\inst{1}\orcidlink{0000-0001-6686-2576} \and
Mufhumudzi Muthivhi \inst{1}\orcidlink{0000-0003-0509-6235} \and
Terence L. van Zyl\inst{1}\orcidlink{0000-0003-4281-630X} \and
Fredrik Gustafsson \inst{2}\orcidlink{0000-0003-3270-171X}}

\authorrunning{J. Huo et al.}

\institute{University of Johannesburg, South Africa 
\email{216045414@student.uj.ac.za, mmuthivhi@uj.ac.za, tvanzyl@uj.ac.za} \and
Linköping University, Sweden
\email{fredrik.gustafsson@liu.se}}
%
%

%
\maketitle              

\begin{abstract}
Current state-of-the-art Wildlife classification models are trained under the closed world setting. When exposed to unknown classes, they remain overconfident in their predictions.
Open-set Recognition (OSR) aims to classify known classes while rejecting unknown samples. Several OSR methods have been proposed to model the closed-set distribution by observing the feature, logit, or softmax probability space. A significant drawback of many existing approaches is the requirement to retrain the pre-trained classification model with the OSR-specific strategy.
This study contributes a post-processing OSR method that measures the agreement between the models' features and predicted logits.
We propose a probability distribution based on an input's distance to its Nearest Class Mean (NCM). The NCM-based distribution is then compared with the softmax probabilities from the logit space to measure agreement between the NCM and the classification head.
Our proposed strategy ranks within the top three on two evaluated datasets, showing consistent performance across the two datasets. In contrast, current state-of-the-art methods excel on a single dataset. We achieve an AUROC of 93.41 and 95.35 for African and Swedish animals.
The code can be found \href{https://github.com/Applied-Representation-Learning-Lab/OSR}{\textbf{here}}.

\end{abstract}
\begin{figure}[htp]
  \centering
  \includegraphics[width=0.95\columnwidth]{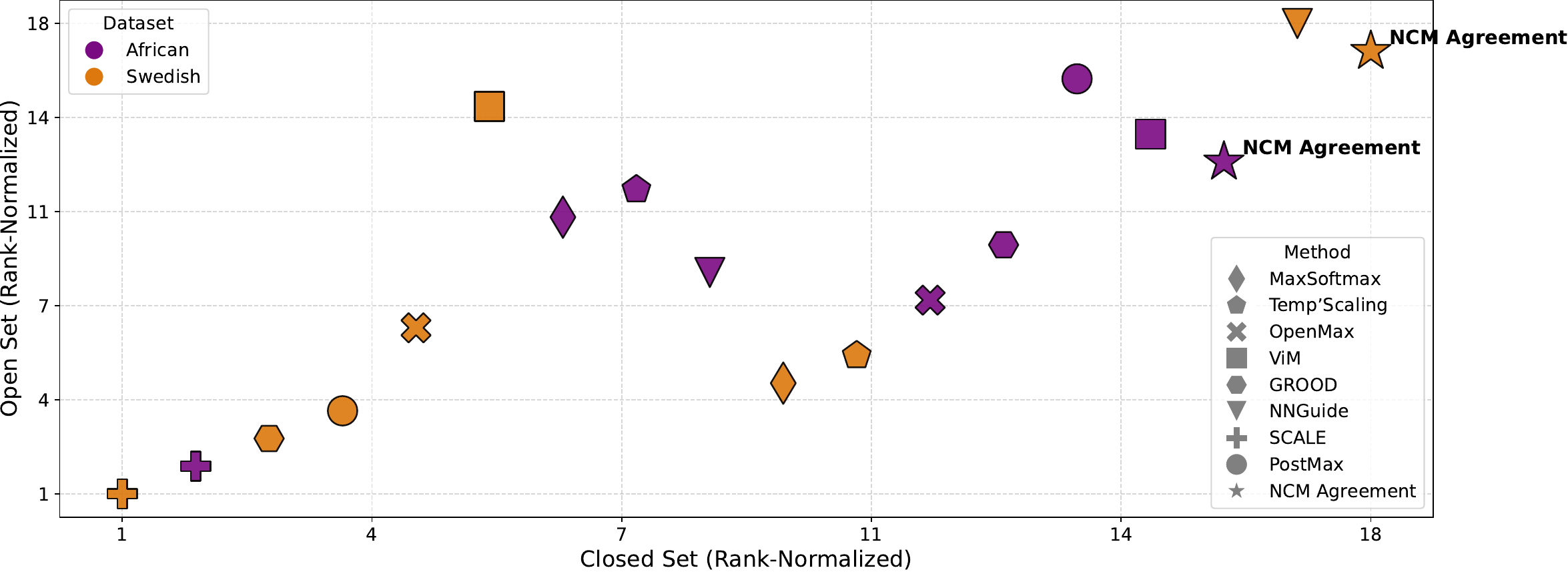} 
  \caption{Closed vs Open set AUROC performance of the different OSR methods across African and Swedish wildlife. Our NCMAgreement strategy (star) consistently produces a high open and closed set performance for both datasets. Other methods achieve optimal performance on one dataset.}\label{figure:openvclosed}
\end{figure}
\section{Introduction}
Wildlife classification models have proven to be useful in wildlife monitoring and ecological studies~\cite{berger2017wildbook}. Several large-scale wildlife classification models have achieved remarkable success over a large variety of animal classes~\cite{villa2017towards, swanson2015snapshot, willi2019identifying, stevens2024bioclip, Beery_Efficient_Pipeline_for, gadot2024crop}. The largest of which, SpeciesNet, can classify up to $2000$ animals.

However, these models are trained under the closed world setting~\cite{yang2024generalized}. They perform well over the classes they have seen during training, but will misclassify unknown classes as known classes. Researchers would have to train the model on every species in a region, ensuring that all possible classes have been seen.

Open-set Recognition (OSR) was proposed to address the limitation of machine learning systems to handle inputs from classes unseen during training \cite{scheirer2014probability, scheirer2012toward}. When an unknown sample is not correctly rejected, it is misclassified as a known class, which in turn reduces overall accuracy. Early OSR research proposed evaluation protocols designed to reflect real-world scenarios, aiming to assess performance more effectively \cite{bendale2016towards, rudd2017extreme}. OSR was meant to improve the reliability of real-world systems. Achieving this requires evaluation protocols and methods that remain robust under varying proportions of unknown classes and align closely with operational needs.  However, several proposed OSR methods require the pre-trained classification model to be re-trained with the OSR-specific strategy~\cite{kong2021opengan, neal2018open, chen2021adversarial}.

This study develops a simple but effective post-processing method for OSR. We study the uncertainty in the models' feature and logit space by measuring their agreement. Experiments are conducted on two datasets from different environments. We compare our approach to several state-of-the-art post-processing OSR and out-of-distribution (OOD) methods.
Our proposed NCMAgreement (Nearest Class Mean Agreement) strategy achieves an AUROC of 93.41 and 95.08 over two datasets. Although our method does not establish state-of-the-art performance on either dataset individually, it demonstrates the most consistent performance across both. Figure~\ref{figure:openvclosed} depicts the performance of our model (the star) against the current state-of-the-art. NCMAgreement achieves a high closed-set accuracy and open-set performance for both datasets. Most methods achieve optimal performance for one dataset.

We contribute to the literature by:
\begin{enumerate}
    \item providing a classification and OSR detection model for African and Swedish animals;
    \item establishing a post-processing OSR method that measures uncertainty within the model's predictions;
    \item a curated dataset of African and Swedish animals for closed and open-set animals;
\end{enumerate}

\section{Background}
Wildlife monitoring is gaining more attention across different environments~\cite{saoud2024beyond}. Villa~\textit{et al.} uses camera-trap images from the Snapshot Serengeti dataset with a neural network to classify species, while uses citizen science to label Serengeti camera-trap data~\cite{villa2017towards, willi2019identifying}. MegaClassifier is trained on cropped MegaDetector images, mostly of North American and European species~\cite{Beery_Efficient_Pipeline_for}. BioClip is trained on the TreeOfLife-10M dataset, which covers many animals, plants, fungi, and insects~\cite{stevens2024bioclip}. SpeciesNet combines MegaDetector with an ensemble model trained on more $60$ million images of about $2,000$ species. However, all of these systems are closed-world models, which means that they cannot identify species that are not in their training data. Open-set recognition (OSR) methods aim to overcome the closed-set limitation of such systems.


\subsection{Related Works}
OSR is different from out-of-distribution (OOD) detection, anomaly detection, and finding new categories. Some think OSR can be solved by first running OOD detection and then performing normal classification. However, such an approach would not always work, especially when the change is in how the image looks, not in what it contains. For example, if we need to recognise known species in night-time infrared images, even though the model was trained only on daylight photos. These infrared images look very different from the training data, but the animals are still known classes and should not be rejected. OSR can be improved in two ways: first training the network to learn stronger and more robust features. Secondly, using post-processing, where a model trained for closed-set classification is adapted for OSR. Our work uses the second approach and makes sure evaluation stays closer to OSR rather than pure OOD detection. We compare several OSR and OOD methods to our approach. Specifically, we explore two thresholding-based methods, MaxSoftmax \cite{hendrycks2017a} and Temperature Scaling \cite{guo2017calibration}. We also consider two OSR post-processing methods, OpenMax \cite{bendale2016towards} and PostMax \cite{cruz2024operational} as well as four OOD methods VIM \cite{wang2022vim}, GROOD \cite{elaraby2023grood} and NNGuide\cite{park2023nearest}, and SCALE \cite{xu2023scaling}.


\subsection{Thresholding Methods}
Thresholding methods are among the simplest and most common approaches for open-set recognition. The main idea is to accept a prediction only when its softmax score or logit value is above a certain threshold. Samples that fall below this value are treated as unknown. These methods require no changes to the model architecture or additional training data. Their performance can be improved by normalizing the logits or applying calibration techniques such as temperature scaling, which help separate known and unknown samples more clearly \cite{hendrycks2017a, guo2017calibration}. Due to their simplicity and generality, thresholding serves as a standard baseline for out-of-distribution detection across many architectures, including Vision Transformers, Masked Autoencoders, and ResNets \cite{cruz2024operational, pmlr-v162-hendrycks22a}. However, thresholding still depends heavily on the confidence levels produced by the pre-trained model. When these confidence scores are poorly calibrated, the model can remain overconfident on unfamiliar data. As a result, threshold-based approaches are best viewed as useful baselines rather than complete solutions for open-set recognition.

\subsection{Post-processing Methods}
Beyond thresholding, post-processing methods aim to refine uncertainty estimation by analyzing the feature space of pre-trained closed-set networks. These methods use the extracted representations to estimate how likely an input belongs to an unknown class instead of retraining the model. PostMax~\cite{cruz2024operational} shows that the magnitude of feature activations tends to differ between known and unknown samples. PostMax models the logits with a Generalized Pareto Distribution and normalizes the resulting scores by feature magnitude to improve separability. NNGuide~\cite{park2023nearest} measures distances in the feature space using $k$-nearest neighbours (KNN) and adjusts the softmax confidence based on how close a sample is to known examples. Other methods such as OpenMax and GROOD \cite{vareto2024open,dhamija2018reducing} model the distribution of features to introduce an “unknown” category. Some recent approaches attach lightweight projection networks to existing feature extractors, allowing them to perform open-set recognition without retraining the full model.

\section{Methodology}
In open-set recognition, classifiers are often overconfident about unseen classes, and softmax scores alone cannot reliably distinguish between known and unknown inputs~\cite{muthivhi2025improving}. Prototype-based methods, such as NCM, capture feature similarity but produce weaker decision boundaries than a trained classifier. Our approach combines these two complementary scores by measuring the agreement between feature–prototype distances and classifier probabilities. When the two scores align, the sample is likely to be known, while disagreement suggests an unknown, resulting in a more robust strategy for open-set detection.

We consider the open-set setting, where test images may belong either to the set of known classes seen during training or to unknown classes. Our goal is to classify known samples and reject unknown samples correctly. We adopt BioClip-2 as our backbone encoder $f$~\cite{gu2025bioclip}. Given an image $x$, we extract its features $z = f(x)$ by freezing $f$. We train a two-layer classification head $g$ separated with a ReLU to obtain a prediction $y \in \mathbb{R}^{n}$, where $n$ is the number of predicted classes.

\begin{figure}[htp]
  \centering
  \includegraphics[width=1.0\columnwidth]{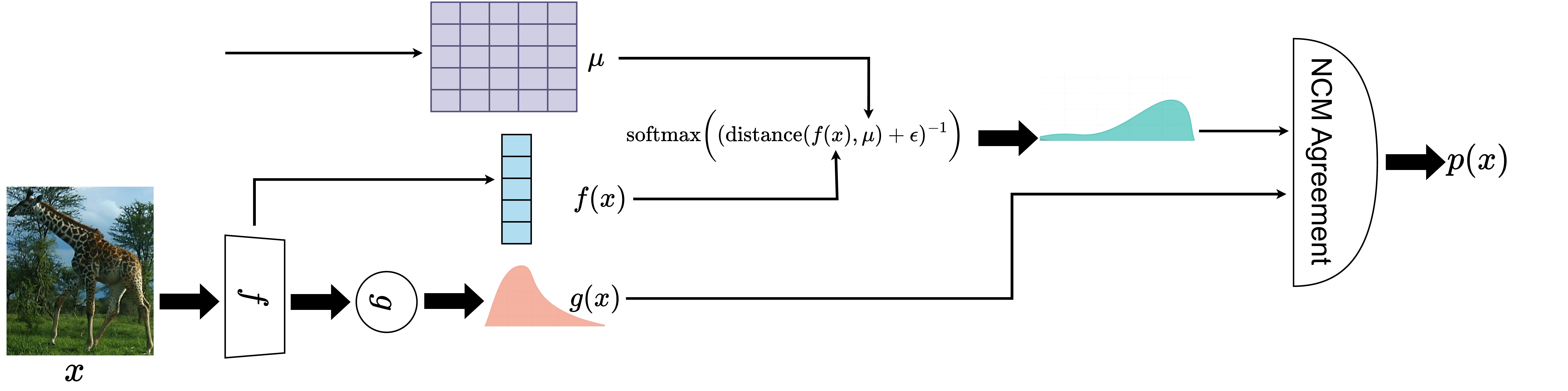} 
\caption{Diagram of our NCMAgreement method, where $f$ denotes the backbone, $g$ denotes the two-layer classifier, and $u$ denotes the mean feature vector for each class. $p$ denotes the NCMAgreement function that takes in both the NCM and classifier scores to output the final probability value between known and unknown inputs.}

  \label{figure:model}
\end{figure}

\subsection{Distance and Prediction Probability Distributions}

To tackle open-set recognition, our method aims to measure the agreement between the pretrained features and the predicted logits. First, we use the Nearest Class Mean (NCM) classifier to extract prototypes from the feature space. For each known class \(c\), we compute its mean feature vector (prototype) from the frozen encoder \(f\) over the validation set \(\mathcal{D}\):
\begin{equation}
    \mu_c = \frac{1}{| \mathcal{D}_c |} \sum_{x \in \mathcal{D}_c} f(x),
\end{equation}
where \(\mu_c \in \mathbb{R}^{d_1}\) and \(d_1\) is the feature dimension of the frozen backbone \(f\). For each image $x$, we calculate the Euclidean distance
\begin{equation}
    v_{c}^{\mathrm{dist}} = \| f(x) - \mu_c \|_2
\end{equation}
such that the $v_{c}^{\mathrm{dist}}$ describes the distance of $x$ to class mean $\mu_c$. We do this for each class and apply an inverse operation over the distances to ensure that higher values correspond to closer proximity to class prototypes. Finally, we apply a $\mathrm{softmax}$ normalization to produce a probability distribution, such that
\begin{equation}
    \mathbf{v}^{\mathrm{dist}} = \mathrm{softmax} \biggl( \left[ \frac{1}{v_{c}^{dist} + \epsilon} \mid c = 1, \dots,n \right] \biggr)
\end{equation}
where $\epsilon \ll 1$ is a small smoothing constant to avoid division by zero.

To decide whether \(x\) belongs to a known or unknown class, we measure the alignment of the distance probability distributions to the $\mathrm{softmax}$ logits produced by the classification head $g$. The result is a vector of class probabilities defined as:
\begin{equation}
    \mathbf{v}^{\mathrm{prob}} = \mathrm{softmax} \Bigl( \left[ g_c(f(x)) \mid c = 1, \dots,n \right] \Bigr).
\end{equation}
$\mathbf{v}^{\mathrm{dist}}$ captures inference-based predictions and $\mathbf{v}^{\mathrm{prob}}$ encodes feature-based information. 

\subsection{NCM Agreement}
We measure the agreement between the distance and prediction distributions to evaluate how much the features and logits agree on the predicted classes. Unlike existing post-processing methods, which estimate uncertainty from only one target such as logits or features, our approach compares the full probability distributions from both the feature and classifier spaces. The method evaluates how consistently these two output distributions rank the known classes instead of relying only on the most confident prediction.  Alignment between the feature-based and classifier-based probabilities indicates agreement, while divergence reflects uncertainty. This consistency in space provides a stronger and more reliable basis for detecting unknown samples.
Given $\mathbf{v}^{\mathrm{dist}}$ and $\mathbf{v}^{\mathrm{prob}}$ we obtain the agreement score
\begin{equation}
    p(\mathbf{v}^{\mathrm{dist}}, \mathbf{v}^{\mathrm{prob}}) = (1 - \mathrm{JS}(\mathbf{v}^{\mathrm{dist}}, \mathbf{v}^{\mathrm{prob}})) \times \Bigl( 1 - \frac{\mathrm{H(\mathbf{v}^{\mathrm{dist}})}}{\log_{2}(n)} \Bigr) \times \Bigl( 1 - \frac{\mathrm{H(\mathbf{v}^{\mathrm{prob}})}}{\log_{2}(n)} \Bigr)
\end{equation}
where $\mathrm{H}$ is the entropy  $\mathrm{JS}$ is Jensen-Shannon divergence given by
\begin{equation}
    \mathrm{H}(p) = - \sum_{i=1}^{n} p_{i} log p_{i}
\end{equation}
and
\begin{equation}
    \mathrm{JS}(v_{1}, v_{2}) = \frac{1}{2} [\mathrm{D}_{\mathrm{KL}}(v_{1} || m) + \mathrm{D}_{\mathrm{KL}}(v_{2} || m)]
\end{equation}
such that $\odot$ is an element-wise multiplication operation, $\mathrm{D}_{\mathrm{KL}}(\cdot || \cdot)$ is the Kullback-Leibler (KL) divergence and $m = \frac{1}{2} (v_1 + v_{2})$ is the midpoint distribution. $\log_{2}(n)$ is the maximum possible entropy for a discrete distribution of size that normalizes the entropy to the range $[0, 1]$. The Jensen-Shannon term measures the similarity of the two vectors. The entropy terms penalize high entropy or uncertainty in both vectors. 

\subsection{Experimental Setup}
We use state-of-the-art post-processing Open-set-Recognition (OSR) and Out-of-Distribution (OOD) benchmarks. The models are obtained from the PyTorch-OOD library~\cite{kirchheim2022pytorch} or their original source directory. We use Pytorch Lightning to streamline our training and testing process~\cite{pytorch2019lightning}. We begin by training a frozen encoder architecture with a classification head over the closed-set datasets. Training is performed using the Weighted Adam optimizer with a learning rate of $0.005$ combined with a cosine warmup schedule. All models are trained for up to $500$ epochs.


\begin{figure}[htp!]
  \centering

  \begin{subfigure}{0.32\textwidth}
    \includegraphics[width=\linewidth]{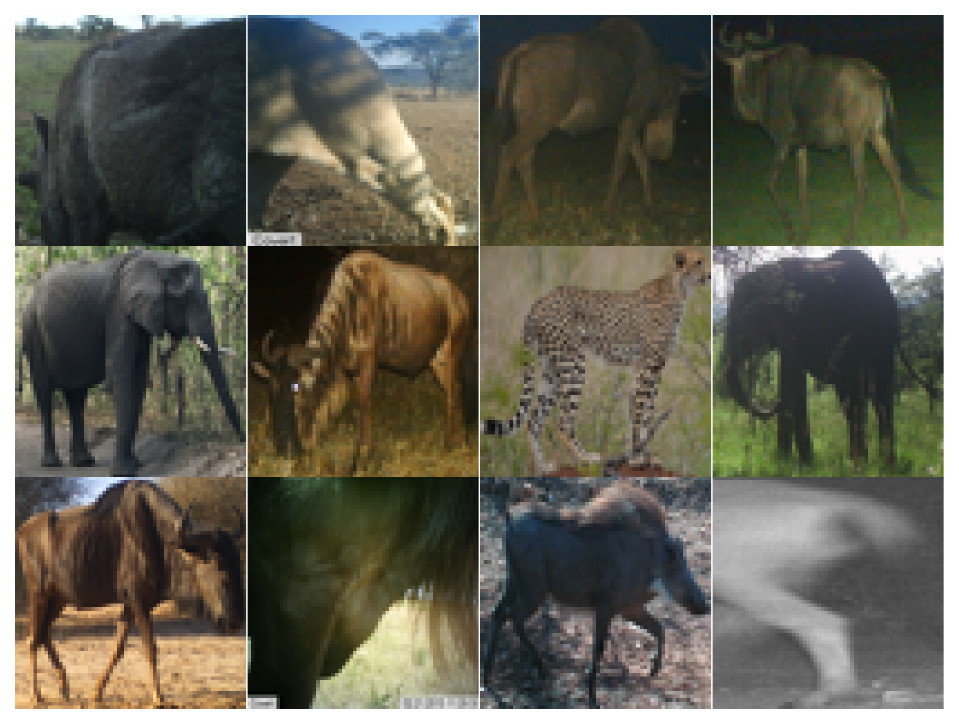}
    \caption{African}
    \label{fig:african}
  \end{subfigure}
  \hfill 
  \begin{subfigure}{0.32\textwidth}
    \includegraphics[width=\linewidth]{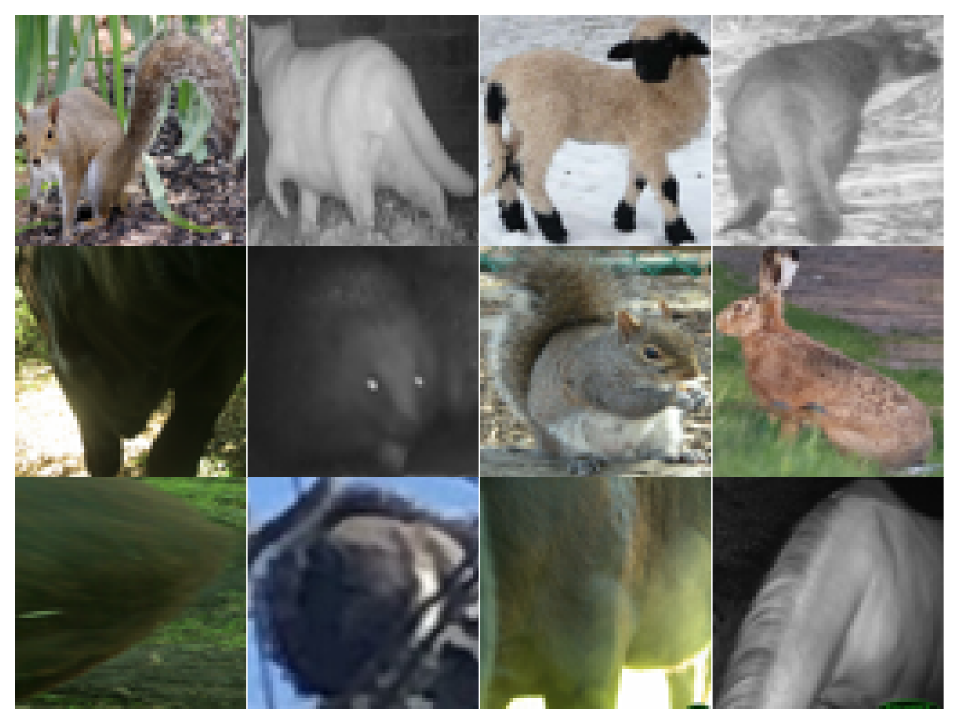}
    \caption{Swedish}
    \label{fig:swedish}
  \end{subfigure}
  \hfill 
  \begin{subfigure}{0.32\textwidth}
    \includegraphics[width=\linewidth]{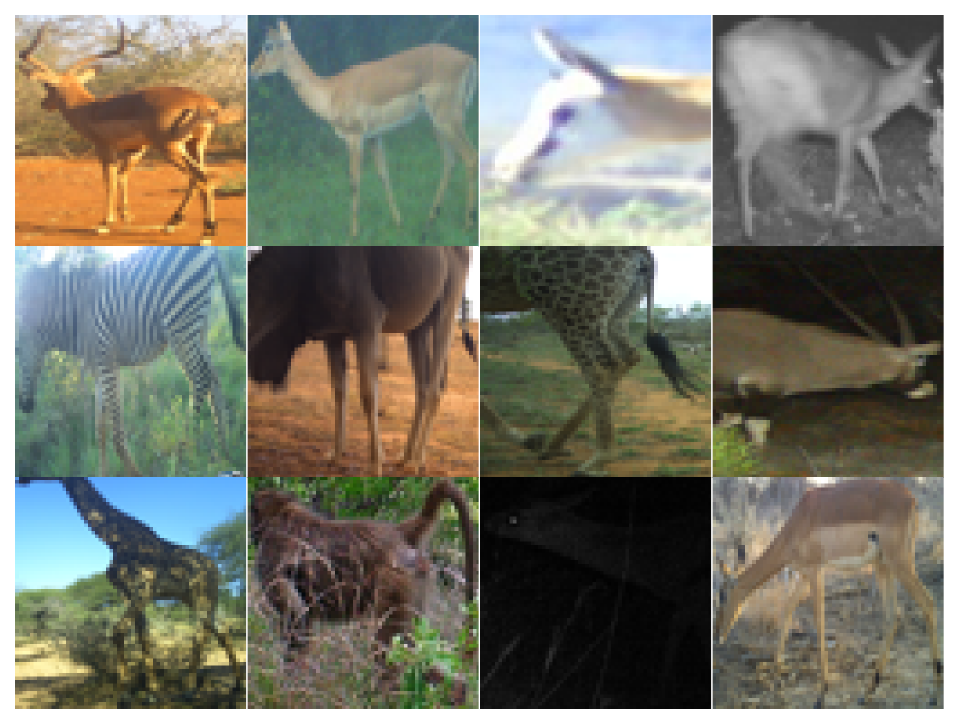}
    \caption{OSR}
    \label{fig:osr}
  \end{subfigure}

  \caption{Examples of images in each of the African, Swedish, and OSR datasets that we use.}
  \label{figure:samples}
\end{figure}

\subsubsection{Dataset}

\begin{table}[htb!]
    \centering
    \caption{Number of images for closed set and open set classes in each of the datasets. (*) Not included during evaluation of models trained on the Swedish dataset}
    \resizebox{0.8\columnwidth}{!}{%
    \begin{tabular}{lr|lr|lr}
        \toprule
        \multicolumn{4}{c|}{Closed Set} & \multicolumn{2}{c}{Open-set} \\
        \multicolumn{2}{c|}{African} & \multicolumn{2}{c|}{Swedish} \\
        Species   & Images \# & Species     & Images \# & Species     & Images \# \\
        \bottomrule \toprule
        Elephant    & $58,517$  & Bear/Wolverine    & $3,562$         & Zebra         & $36,219$ \\
        Buffalo     & $16,660$   & Bird              & $83,095$      & Gazelle       & $32,499$ \\
        Rhino       & $8,378$   & Cat               & $97,771$      & Implala       & $51,891$ \\
        Lion        & $15,932$   & Cattle            & $567,679$     & Giraffe       & $21,527$ \\
        Felidae     & $19,489$   & Deer              & $164,210$      & Baboon        & $11,977$ \\
        Canine      & $96,890$  & Dog               & $3,957$         & Hartebeest    & $3,881$ \\
        Hyena       & $19,995$   & Fowl              & $4,684$       & Gemsbok       & $19,437$ \\
        Hippo       & $8,419$   & Fox               & $42,347$      & Eland         & $11,162$ \\
        Wildpig     & $24,771$   & Horse             & $16,017$       & Bovine*        & $16,828$ \\
        Wildebeest  & $50,248$   & Livestock         & $8,972$       & Skunk*         & $52$ \\
                    &           & Moose             & $16,284$       & Bontebok      & $4,147$ \\
                    &           & Mustelidae        & $66,043$      & Ostriche      & $4,613$ \\
                    &           & Rabbit/Hare       & $39,132$       & Duiker        & $8,428$ \\
                    &           & Raccoon           & $51,580$      & Monkey        & $2,677$ \\
                    &           & Rodent            & $129,187$     & Steenbok      & $6,370$ \\
                    &           & Snake             & $24,803$        & Dik dik       & $4,023$ \\
                    &           & Sus               & $73,686$      & Hare*          & $2,110$ \\
                    &           & Wolf              & $1,658$         & Porcupine     & $1,204$ \\
                    &           &                   &               & Antelope      & $1,695$ \\
                    &           &                   &               & Springbok     & $14,109$  \\
                    &           &                   &               & Mongoose      & $623$  \\
                    &           &                   &               & Nyala         & $853$  \\
                    &           &                   &               & Duiker        & $8,428$  \\
                    &           &                   &               & Aardvark      & $938$  \\
                    &           &                   &               & Badger*        & $148$  \\
                    &           &                   &               & Tortoise*      & $182$  \\
        \midrule
        Training    &  $319,299$    & Training      & $1,394,667$   &                &  \\
        Test        &  $79,820$     & Test          & $348,655$     & Test          & $268,957$ \\
        \bottomrule
    \end{tabular}
    }    
    \label{tab:animal_images}
\end{table} 

To evaluate OSR in real-world settings, we construct a dataset based on wildlife imagery from the LILA BC repositories and TreeOfLife~\cite{lila2024science, gu2025bioclip}. We focus on species relevant to South African and Swedish wildlife. Our selection of these regions is motivated by the opportunity to evaluate the model’s effectiveness across two distinct ecosystems, sub-Saharan Africa and Northern Europe. We use MegaDetector to crop animals from camera trap images, applying a confidence threshold of 0.5. To address data leakage between splits, K-means clustering is applied to group similar images, ensuring distinct clusters across training, validation, and test sets. We only consider images that have a single animal in them to avoid mislabeling background animals. Dark and bright images were filtered out by considering the average pixel value. We cluster similar images together by placing them all in the training set, while the testing set only consists of non-duplicates~\cite{adam2024wildlifereid}. We first extract the features of each image with DINOv2~\cite{oquab2023dinov2}. Then HDBScan automatically clusters similar features together~\cite{mcinnes2017hdbscan}. We sample one image in each cluster to produce the test set. The splits were stratified by class labels to preserve the ratios between classes.

In addition to the closed-set species, we randomly select visually similar species to serve as unknown classes for open-set evaluation. These open-set species are deliberately chosen to be similarly close to closed-set classes, increasing the difficulty of the recognition task. For instance, cheetahs and giraffes share spotted textures with leopards under certain lighting, while impalas, wildebeests, and hippos may resemble buffalo, lions, or rhinos in terms of silhouette and pose. Elephants and zebras often share habitats with other species, leading to potential overlap between the classes, thus increasing the OSR difficulty.

Table~\ref{tab:animal_images} summarizes the number of images per species and their distribution across close and open sets. We split each dataset into training and test sets by stratifying the class labels to preserve their ratios. All images are resized and center-cropped to \(224 \times 224\) before being passed to the model.

\subsubsection{Metrics}
\label{sec:metrics}
We adopt four commonly used metrics for evaluating OSR performance. The Area Under the Receiver Operating Characteristic Curve (AUROC) quantifies a model’s ability to distinguish between known (closed-set) and unknown (open-set) classes at varying decision thresholds~\cite{fawcett2006introduction}. An AUROC of $1$ indicates perfect separation between closed and open sets, while $0.5$ is random guessing. The Area Under the Precision-Recall Curve (AUPR) captures the balance between precision and recall~\cite{powers2020evaluation}; AUPR-IN treats closed-set samples as positives, while AUPR-OUT considers open-set samples as positive classes, making it suitable for OSR contexts where the detection of unknowns is critical. The F1-score evaluates how a model balances its precision and recall performance under one single metric. A value of one means that the model has correctly classified all candidates. We present the macro and weighted F1-score. Finally, we have the AUROC difference to show the absolute difference between the AUROC values of the African and Swedish models on the OSR dataset. The value tells us the OSR performance stability of each of the OSR/OOD methods across different models.

\section{Results}

\begin{table}[htb!]
\caption{Open-set recognition performance comparison across African and Swedish wildlife datasets using AUROC, FPR95-TPR, AUPR-IN, and AUPR-OUT metrics. Lower AUROC difference indicates more consistent cross-dataset performance. Highlighted and bold numbers indicate the best performance, while highlighted numbers indicate the second-best performance.}
\centering
\resizebox{\textwidth}{!}{%
\begin{tabular}{lrrrr|rrrrr}
\toprule
\textit{\textbf{Metrics}} & 
    \multicolumn{4}{c}{\textbf{African}} & 
    \multicolumn{4}{c}{\textbf{Swedish}} \\
\cmidrule(lr){2-5} \cmidrule(lr){6-9} \\
\textit{\textbf{Models}} 
& \makecell[l]{AUROC $\uparrow$} & \makecell[l]{FPR95-\\TPR $\downarrow$} & \makecell[l]{AUPR-\\IN $\uparrow$} & \makecell[l]{AUPR-\\OUT $\uparrow$}
& \makecell[l]{AUROC $\uparrow$} & \makecell[l]{FPR95-\\TPR $\downarrow$} & \makecell[l]{AUPR-\\IN $\uparrow$} & \makecell[l]{AUPR-\\OUT $\uparrow$} & \makecell[l]{AUROC\\Difference $\downarrow$} \\
\bottomrule
\toprule
\; MaxSoftmax (ICLR’17~\cite{hendrycks2017a})  & \makecell[l]{93.13} & \makecell[l]{19.22} & \makecell[l]{89.21} & \makecell[l]{96.64} & \makecell[l]{91.22} & \makecell[l]{28.86} & \makecell[l]{94.42} & \makecell[l]{83.64} & \makecell[l]{2.25} \\
\; Temp'Scaling (PMLR'17~\cite{guo2017calibration}) & \makecell[l]{93.15} & \makecell[l]{19.18} & \makecell[l]{89.24} & \makecell[l]{96.65} & \makecell[l]{91.24} & \makecell[l]{28.78} & \makecell[l]{94.43} & \makecell[l]{83.66} & \makecell[l]{1.91} \\
\; OpenMax (CVPR'16~\cite{bendale2016towards})     & \makecell[l]{92.50} & \makecell[l]{24.53} & \makecell[l]{85.51} & \makecell[l]{96.44} & \makecell[l]{91.61} & \makecell[l]{21.46} & \makecell[l]{95.33} & \makecell[l]{82.08} & \ccB{\makecell[l]{1.91}} \\
\; VIM (CVPR’22~\cite{wang2022vim})        & \ccb{\makecell[l]{93.52}} & \ccB{\makecell[l]{14.36}} & \ccB{\makecell[l]{91.71}} & \makecell[l]{96.43} & \ccb{\makecell[l]{94.27}} & \ccb{\makecell[l]{16.24}} & \ccb{\makecell[l]{96.72}} & \ccb{\makecell[l]{90.34}} & \ccB{\makecell[l]{0.75}} \\
\; GROOD (ICCVW'23~\cite{vojivr2023calibrated})         & \makecell[l]{93.01} & \makecell[l]{22.43} & \makecell[l]{88.23} & \ccb{\makecell[l]{96.76}} & \makecell[l]{75.08} & \makecell[l]{62.86} & \makecell[l]{86.31} & \makecell[l]{72.57} & \makecell[l]{17.03} \\
\; NNGuide (ICCV’23~\cite{park2023nearest})        & \makecell[l]{92.85} & \makecell[l]{28.67} & \makecell[l]{86.56} & \ccb{\makecell[l]{97.06}} & \ccB{\makecell[l]{97.82}} & \ccB{\makecell[l]{8.81}} & \ccB{\makecell[l]{98.53}} & \ccB{\makecell[l]{96.43}} & \makecell[l]{4.97} \\
\; SCALE (ICLR’24~\cite{xu2023scaling})       & \makecell[l]{60.70} & \makecell[l]{85.76} & \makecell[l]{34.16} & \makecell[l]{81.28} & \makecell[l]{45.76} & \makecell[l]{97.16} & \makecell[l]{54.71} & \makecell[l]{38.61} & \makecell[l]{14.94} \\
\; PostMax (CVPR’24~\cite{cruz2024operational})        & \ccB{\makecell[l]{94.21}} & \ccb{\makecell[l]{18.29}} & \ccb{\makecell[l]{89.45}} & \ccB{\makecell[l]{97.34}} & \makecell[l]{80.62} & \makecell[l]{58.49} & \makecell[l]{86.36} & \makecell[l]{71.71} & \makecell[l]{13.59} \\
\; \textbf{NCM Agreement Score}         & \ccb{\makecell[l]{93.41}} & \ccb{\makecell[l]{14.85}} & \ccb{\makecell[l]{91.27}} & \makecell[l]{96.07} & \ccb{\makecell[l]{95.35}} & \ccb{\makecell[l]{16.66}} & \ccb{\makecell[l]{97.09}} & \ccb{\makecell[l]{91.26}} & \ccb{\makecell[l]{1.94}} \\
\bottomrule
\end{tabular}
}
\label{table:OSR}
\end{table}

\subsection{Open-Set Recognition}

\begin{table}[htbp!]
\caption{Per-species accuracy and average F1 scores for closed-set and open-set recognition methods on African and Swedish wildlife datasets}
\label{table:accuracy}
\centering
\resizebox{0.8\columnwidth}{!}{
\begin{tabular}{rlr|rrrrrrrrrr}
\toprule
\rot{\makecell[l]{\textit{\textbf{Species}}}}
& & \rot{\makecell[l]{Closed-\\set}} &\rot{\makecell[l]{Max-\\Softmax}} & \rot{\makecell[l]{Temp'-\\Scaling}} & \rot{\makecell[l]{OpenMax}} & \rot{\makecell[l]{VIM}} & \rot{\makecell[l]{GROOD}} & \rot{\makecell[l]{NNGuide}} & \rot{\makecell[l]{SCALE}} & \rot{\makecell[l]{PostMax}} & \rot{\makecell[l]{NCM-\\Agreement}} \\
\midrule
\toprule
\multirow{12}{*}{\rot{African}} & {Elephant}          & \makecell[l]{97.04} & \makecell[l]{84.91} & \makecell[l]{84.95} & \makecell[l]{84.58} & \makecell[l]{84.70} & \makecell[l]{82.90} & \makecell[l]{87.03} & \makecell[l]{19.87} & \makecell[l]{86.66} & \makecell[l]{85.95} \\ 
& {Buffalo}           & \makecell[l]{86.43} & \makecell[l]{58.27} & \makecell[l]{58.32} & \makecell[l]{70.76} & \makecell[l]{74.62} & \makecell[l]{76.18} & \makecell[l]{68.50} & \makecell[l]{27.88} & \makecell[l]{70.08} & \makecell[l]{78.15} \\ 
& {Rhino}             & \makecell[l]{88.78} & \makecell[l]{64.66} & \makecell[l]{64.90} & \makecell[l]{74.07} & \makecell[l]{74.16} & \makecell[l]{78.75} & \makecell[l]{74.07} & \makecell[l]{40.26} & \makecell[l]{75.12} & \makecell[l]{77.55} \\ 
& {Lion}              & \makecell[l]{92.11} & \makecell[l]{78.30} & \makecell[l]{78.30} & \makecell[l]{76.29} & \makecell[l]{78.60} & \makecell[l]{79.58} & \makecell[l]{78.30} & \makecell[l]{44.78} & \makecell[l]{73.13} & \makecell[l]{80.39} \\ 
& {Felidae}           & \makecell[l]{97.11} & \makecell[l]{92.30} & \makecell[l]{92.32} & \makecell[l]{92.39} & \makecell[l]{90.27} & \makecell[l]{81.94} & \makecell[l]{88.16} & \makecell[l]{05.36} & \makecell[l]{94.54} & \makecell[l]{87.03} \\ 
& {Canine}            & \makecell[l]{99.09} & \makecell[l]{93.15} & \makecell[l]{93.18} & \makecell[l]{89.46} & \makecell[l]{88.78} & \makecell[l]{77.33} & \makecell[l]{87.48} & \makecell[l]{74.52} & \makecell[l]{92.04} & \makecell[l]{85.27} \\ 
& {Hyena}             & \makecell[l]{93.08} & \makecell[l]{73.21} & \makecell[l]{73.27} & \makecell[l]{73.91} & \makecell[l]{76.89} & \makecell[l]{79.71} & \makecell[l]{70.39} & \makecell[l]{30.93} & \makecell[l]{74.69} & \makecell[l]{73.31} \\ 
& {Hippo}             & \makecell[l]{95.53} & \makecell[l]{82.89} & \makecell[l]{82.89} & \makecell[l]{86.03} & \makecell[l]{83.79} & \makecell[l]{84.03} & \makecell[l]{83.70} & \makecell[l]{70.20} & \makecell[l]{84.17} & \makecell[l]{86.41} \\ 
& {Wildpig}           & \makecell[l]{95.93} & \makecell[l]{82.98} & \makecell[l]{83.01} & \makecell[l]{82.77} & \makecell[l]{85.59} & \makecell[l]{83.70} & \makecell[l]{78.57} & \makecell[l]{78.70} & \makecell[l]{84.79} & \makecell[l]{84.85} \\ 
& {Wildebeest}        & \makecell[l]{94.77} & \makecell[l]{70.75} & \makecell[l]{70.89} & \makecell[l]{77.14} & \makecell[l]{77.80} & \makecell[l]{83.63} & \makecell[l]{75.48} & \makecell[l]{66.48} & \makecell[l]{77.61} & \makecell[l]{81.68} \\ 
& {Open-set Animals}        & \makecell[l]{-} & \makecell[l]{93.52} & \makecell[l]{93.49} & \makecell[l]{89.30} & \makecell[l]{96.17} & \makecell[l]{91.86} & \makecell[l]{89.62} & \makecell[l]{64.24} & \makecell[l]{92.38} & \makecell[l]{95.54} \\ 
\cmidrule(lr{1em}){2-12}
& \textbf{F1 Score \tiny{(Macro Ave.)}}        & \makecell[l]{94.36} & \makecell[l]{83.21} & \makecell[l]{83.24} & \makecell[l]{79.23} & \makecell[l]{85.85} & \makecell[l]{83.99} & \makecell[l]{79.75} & \makecell[l]{40.21} & \makecell[l]{83.55} & \makecell[l]{85.39} \\ 
& \textbf{F1 Score \tiny{(Weighted Ave.)}}        & \makecell[l]{95.92} & \makecell[l]{91.26} & \makecell[l]{91.26} & \makecell[l]{88.89} & \makecell[l]{93.45} & \makecell[l]{90.36} & \makecell[l]{88.86} & \makecell[l]{64.35} & \makecell[l]{91.12} & \makecell[l]{92.97} \\ 
\midrule
\midrule
\multirow{20}{*}{\rot{Swedish}} & {Bear/Wolverine}  & \makecell[l]{85.42} & \makecell[l]{85.51} & \makecell[l]{85.51} & \makecell[l]{40.90} & \makecell[l]{75.39} & \makecell[l]{85.84} & \makecell[l]{87.08} & \makecell[l]{63.00} & \makecell[l]{57.98} & \makecell[l]{85.96} \\ 
& {Bird}   & \makecell[l]{99.27} & \makecell[l]{94.75} & \makecell[l]{94.76} & \makecell[l]{86.14} & \makecell[l]{78.61} & \makecell[l]{87.81} & \makecell[l]{85.81} & \makecell[l]{30.07} & \makecell[l]{81.59} & \makecell[l]{81.22} \\ 
&  {Cat}     & \makecell[l]{96.95} & \makecell[l]{79.13} & \makecell[l]{79.17} & \makecell[l]{68.15} & \makecell[l]{73.62} & \makecell[l]{45.55} & \makecell[l]{86.26} & \makecell[l]{69.24} & \makecell[l]{52.87} & \makecell[l]{87.22} \\ 
&  {Cattle}      & \makecell[l]{98.95} & \makecell[l]{79.39} & \makecell[l]{79.53} & \makecell[l]{86.47} & \makecell[l]{89.44} & \makecell[l]{09.70} & \makecell[l]{90.51} & \makecell[l]{32.75} & \makecell[l]{85.21} & \makecell[l]{89.87} \\ 
&  {Deer}   & \makecell[l]{97.90} & \makecell[l]{82.04} & \makecell[l]{83.07} & \makecell[l]{77.16} & \makecell[l]{85.92} & \makecell[l]{45.48} & \makecell[l]{92.24} & \makecell[l]{27.11} & \makecell[l]{42.33} & \makecell[l]{87.83} \\ 
&  {Dog}     & \makecell[l]{85.17} & \makecell[l]{59.86} & \makecell[l]{59.96} & \makecell[l]{37.41} & \makecell[l]{36.20} & \makecell[l]{14.76} & \makecell[l]{57.13} & \makecell[l]{43.38} & \makecell[l]{51.37} & \makecell[l]{82.10} \\ 
& {Fowl}     & \makecell[l]{92.52} & \makecell[l]{89.74} & \makecell[l]{89.83} & \makecell[l]{77.18} & \makecell[l]{73.59} & \makecell[l]{66.50} & \makecell[l]{84.62} & \makecell[l]{13.08} & \makecell[l]{61.37} & \makecell[l]{86.92} \\ 
&  {Fox}     & \makecell[l]{96.47} & \makecell[l]{81.59} & \makecell[l]{81.59} & \makecell[l]{56.63} & \makecell[l]{81.31} & \makecell[l]{66.02} & \makecell[l]{90.95} & \makecell[l]{54.37} & \makecell[l]{28.68} & \makecell[l]{89.91} \\ 
&  {Horse}     & \makecell[l]{91.64} & \makecell[l]{68.66} & \makecell[l]{68.66} & \makecell[l]{68.03} & \makecell[l]{74.48} & \makecell[l]{00.00} & \makecell[l]{81.34} & \makecell[l]{29.92} & \makecell[l]{68.11} & \makecell[l]{81.27} \\ 
& {Livestock}     & \makecell[l]{80.61} & \makecell[l]{71.03} & \makecell[l]{72.12} & \makecell[l]{72.57} & \makecell[l]{61.37} & \makecell[l]{19.67} & \makecell[l]{61.73} & \makecell[l]{82.02} & \makecell[l]{74.58} & \makecell[l]{73.06} \\ 
&  {Moose}  & \makecell[l]{92.50} & \makecell[l]{79.75} & \makecell[l]{79.78} & \makecell[l]{76.17} & \makecell[l]{84.28} & \makecell[l]{54.47} & \makecell[l]{89.63} & \makecell[l]{51.29} & \makecell[l]{53.12} & \makecell[l]{89.73} \\ 
& {Mustelidae}   & \makecell[l]{96.26} & \makecell[l]{82.70} & \makecell[l]{82.71} & \makecell[l]{72.16} & \makecell[l]{77.95} & \makecell[l]{65.58} & \makecell[l]{86.32} & \makecell[l]{27.74} & \makecell[l]{59.72} & \makecell[l]{79.39} \\ 
& {Rabbit/Hare}     & \makecell[l]{94.69} & \makecell[l]{89.86} & \makecell[l]{89.88} & \makecell[l]{84.84} & \makecell[l]{83.79} & \makecell[l]{85.36} & \makecell[l]{91.92} & \makecell[l]{34.98} & \makecell[l]{82.69} & \makecell[l]{91.43} \\ 
& {Raccoon}      & \makecell[l]{94.24} & \makecell[l]{76.53} & \makecell[l]{76.56} & \makecell[l]{70.95} & \makecell[l]{80.92} & \makecell[l]{36.78} & \makecell[l]{90.42} & \makecell[l]{42.62} & \makecell[l]{19.77} & \makecell[l]{87.16} \\ 
& {Rodent}   & \makecell[l]{98.41} & \makecell[l]{90.81} & \makecell[l]{90.81} & \makecell[l]{88.60} & \makecell[l]{84.88} & \makecell[l]{82.05} & \makecell[l]{92.01} & \makecell[l]{20.18} & \makecell[l]{72.81} & \makecell[l]{87.38} \\ 
& {Snake}     & \makecell[l]{99.81} & \makecell[l]{99.05} & \makecell[l]{99.05} & \makecell[l]{92.42} & \makecell[l]{88.65} & \makecell[l]{91.37} & \makecell[l]{81.16} & \makecell[l]{24.24} & \makecell[l]{98.95} & \makecell[l]{87.35} \\ 
& {Sus}     & \makecell[l]{92.12} & \makecell[l]{61.57} & \makecell[l]{61.66} & \makecell[l]{68.35} & \makecell[l]{77.72} & \makecell[l]{00.00} & \makecell[l]{83.14} & \makecell[l]{15.68} & \makecell[l]{59.73} & \makecell[l]{83.82} \\ 
& {Wolf}     & \makecell[l]{91.06} & \makecell[l]{66.91} & \makecell[l]{66.91} & \makecell[l]{14.01} & \makecell[l]{56.97} & \makecell[l]{00.00} & \makecell[l]{83.33} & \makecell[l]{57.00} & \makecell[l]{35.51} & \makecell[l]{84.78} \\ 
& {Open-set Animals}        & \makecell[l]{-} & \makecell[l]{89.04} & \makecell[l]{89.02} & \makecell[l]{93.29} & \makecell[l]{93.06} & \makecell[l]{99.19} & \makecell[l]{95.83} & \makecell[l]{70.72} & \makecell[l]{75.69} & \makecell[l]{91.69} \\ 
\cmidrule(lr{1em}){2-12}
& \textbf{F1 Score \tiny{(Macro Ave.)}}        & \makecell[l]{95.44} & \makecell[l]{85.83} & \makecell[l]{85.84} & \makecell[l]{77.39} & \makecell[l]{83.89} & \makecell[l]{57.57} & \makecell[l]{89.07} & \makecell[l]{43.78} & \makecell[l]{66.99} & \makecell[l]{87.36} \\ 
& \textbf{F1 Score \tiny{(Weighted Ave.)}}        & \makecell[l]{97.34} & \makecell[l]{84.88} & \makecell[l]{84.91} & \makecell[l]{85.82} & \makecell[l]{88.45} & \makecell[l]{57.12} & \makecell[l]{92.07} & \makecell[l]{46.30} & \makecell[l]{72.44} & \makecell[l]{88.78} \\ 
\bottomrule
\end{tabular}
}

\end{table}

Table~\ref{table:OSR} presents the Open-set Recognition performance for each baseline method and our proposed NCMAgreement method. The performance of models on the Swedish dataset is higher than on the African dataset. The best AUROC on the Swedish dataset is $97.82$, compared to $94.21$ on the African dataset. Although the OSR samples used all belong to distinct classes, they are still derived from the same dataset as the African dataset. Hence, the OSR samples remain similar to the African samples, providing a more challenging test.

NCMAgreement demonstrates strong performance, achieving the third and second highest AUROC of $93.41$ and $95.08$ on the African and Swedish datasets, respectively. Notably, our proposed approach exhibits consistency across the two datasets, yielding the second-best absolute AUROC difference of $1.67$. This smaller difference indicates that the model's performance remains consistent regardless of the dataset it's trained on.

While Postmax and NNGuide achieve the highest AUROC scores on the African and Swedish datasets, respectively, their performance generalizes less effectively. The results show a high AUROC difference of $13.59$ for Postmax and $4.97$ for NNGuide. PostMax uses a single logit value of the target class and its feature representations to produce an OSR score. Its derived Pareto distribution remains relevant over samples obtained from the same distribution. NNGuide considers the closed-set neighbours of the target class to guide the classifier's output to enforce the boundary geometry of the data manifold. Hence, it performs best when the closed set samples are highly distinguishable from the OSR samples, such as in the Swedish dataset.

Our approach aims to measure uncertainty within the model. The pretrained features and the predicted logits are compared against each other to obtain a measurement of agreement. Hence, the proposed strategy remains fairly consistent across each dataset.

\begin{figure}[htp]
  \centering

  \includegraphics[width=0.95\columnwidth]{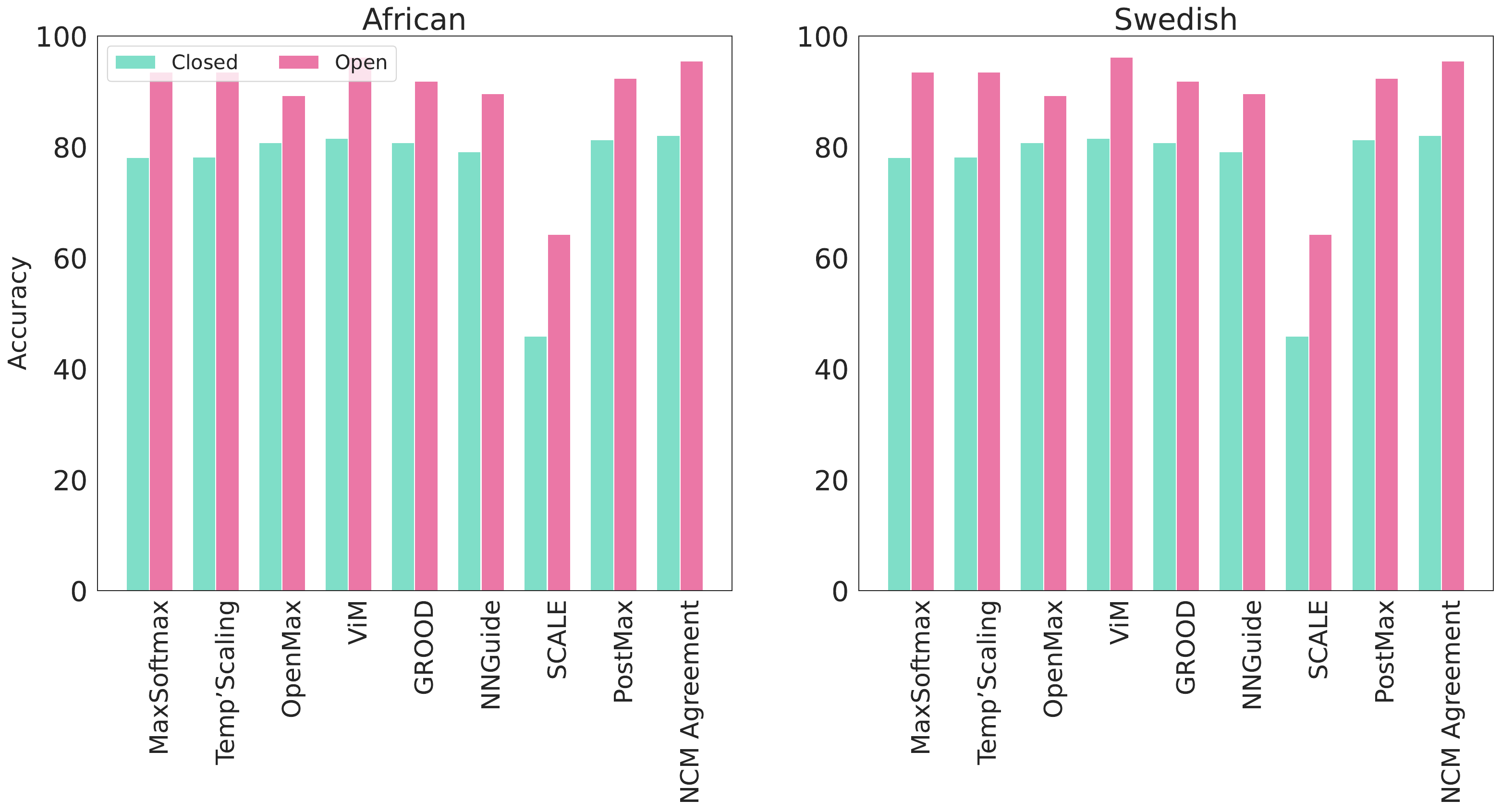} 
      \caption{The F1 score comparison with each of the methods.}\label{figure:f1score}
\end{figure}

\subsection{Closed and open-set Accuracy}

Table~\ref{table:accuracy} and Figure~\ref{figure:f1score} show the closed and open-set accuracy and F1-score achieved by each OSR method. Most models achieve around 90\% accuracy on the open-set. However, these models perform slightly below NCM agreement on the closed-set. Our proposed method produces the highest closed-set accuracy while achieving second and third place on the open-set for the African and Swedish datasets, respectively. Notably, ViM displays the same high closed and open-set accuracy. It aims to simulate the logit of an open-set sample using the feature space and predicted logits of the closed-set. The premise is that a sample is more likely to be outside the training distribution if it has a smaller original logit value and a larger residual of its feature vector against a principal subspace. Our NCM agreement strategy addresses a similar pattern within the feature and logit spaces by measuring the alignment between the two spaces to determine the model's uncertainty.

\section{Discussion}
The results of this study suggest that current Open-Set Recognition and Out-of-Distribution methods often lack consistent generalization capabilities. The likely reason is that these methods immediately model a closed-set distribution from features, logits, and/or softmax probabilities, assuming the model's outputs remain consistent across each of these output spaces.
Our proposed approach aims to identify disagreement and quantify a measure of uncertainty. 

\section{Conclusion}
This study aims to develop an uncertainty measure for a model’s predictions by evaluating the agreement between two prediction heads. We construct a probability distribution based on an input’s distance to its Nearest Class Mean and then quantify the agreement between this NCM-based distribution and the softmax probabilities produced by the classification head.
Although our proposed strategy does not beat current state-of-the-art, its performance remains consistent across datasets under simple and challenging settings.
Future research can look into the OSR of herd detection of animals instead of individual animal detection when animals are heavily occluded by one another.

\section{Acknowledgements}
We would like to thank Qulinda AB and Linkoping
University, Sweden, for providing some of the data and machine-learning equipment. The work was funded by the World Wide Fund for Nature (WWF) with additional support
from the Center for Artificial Intelligence Research (CAIR).

\bibliographystyle{splncs04}
\bibliography{main}
\end{document}